\documentclass[10pt,twocolumn,letterpaper]{article}

\usepackage{cvpr}
\usepackage{times}
\usepackage{epsfig}
\usepackage{graphicx}
\usepackage{amsmath}
\usepackage{amssymb}
\usepackage{bbm}
\usepackage{algorithmic}
\usepackage{wrapfig}
\usepackage{booktabs}
\usepackage{subcaption}
\usepackage{pifont}
\newcommand{\cmark}{\ding{51}}%
\usepackage{multirow}
\usepackage{authblk}

\makeatletter
\newcommand{\printfnsymbol}[1]{%
  \textsuperscript{\@fnsymbol{#1}}%
}
\makeatother

\usepackage[breaklinks=true,bookmarks=false]{hyperref}

\cvprfinalcopy 

\def\cvprPaperID{****} 

\begin{document}

\title{Large-Scale Object Detection in the Wild\\ from Imbalanced Multi-Labels}


\author[1,2,3]{Junran Peng\thanks{Equal contributions.}}
\author[2]{Xingyuan Bu\printfnsymbol{1}}
\author[2]{Ming Sun}
\author[1,3]{Zhaoxiang Zhang\thanks{Corresponding author.(zhaoxiang.zhang@ia.ac.cn)}}
\author[1,3]{Tieniu Tan}
\author[2]{Junjie Yan}
\affil[1]{University of Chinese Academy of Sciences}
\affil[2]{SenseTime Group Limited}
\affil[3]{Center for Research on Intelligent Perception and Computing, CASIA}

\maketitle


    


\begin{figure*}[t]
\begin{subfigure}[b]{0.246\textwidth}
\includegraphics[height=0.140\textheight]{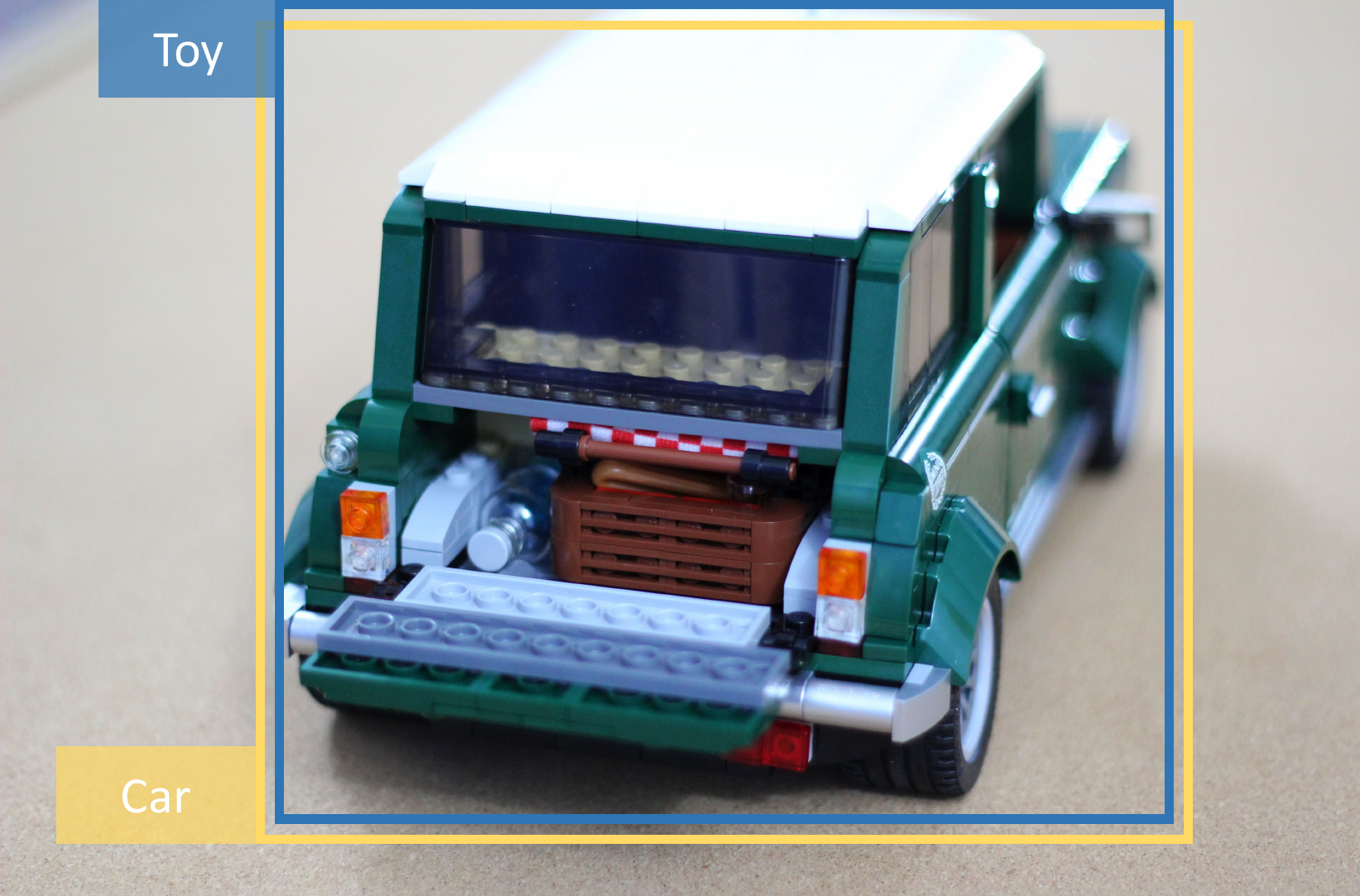}
\caption{}
\label{fig:problem_ex_leaf}
\end{subfigure}
\begin{subfigure}[b]{0.247\textwidth}
\centering
\includegraphics[height=0.140\textheight]{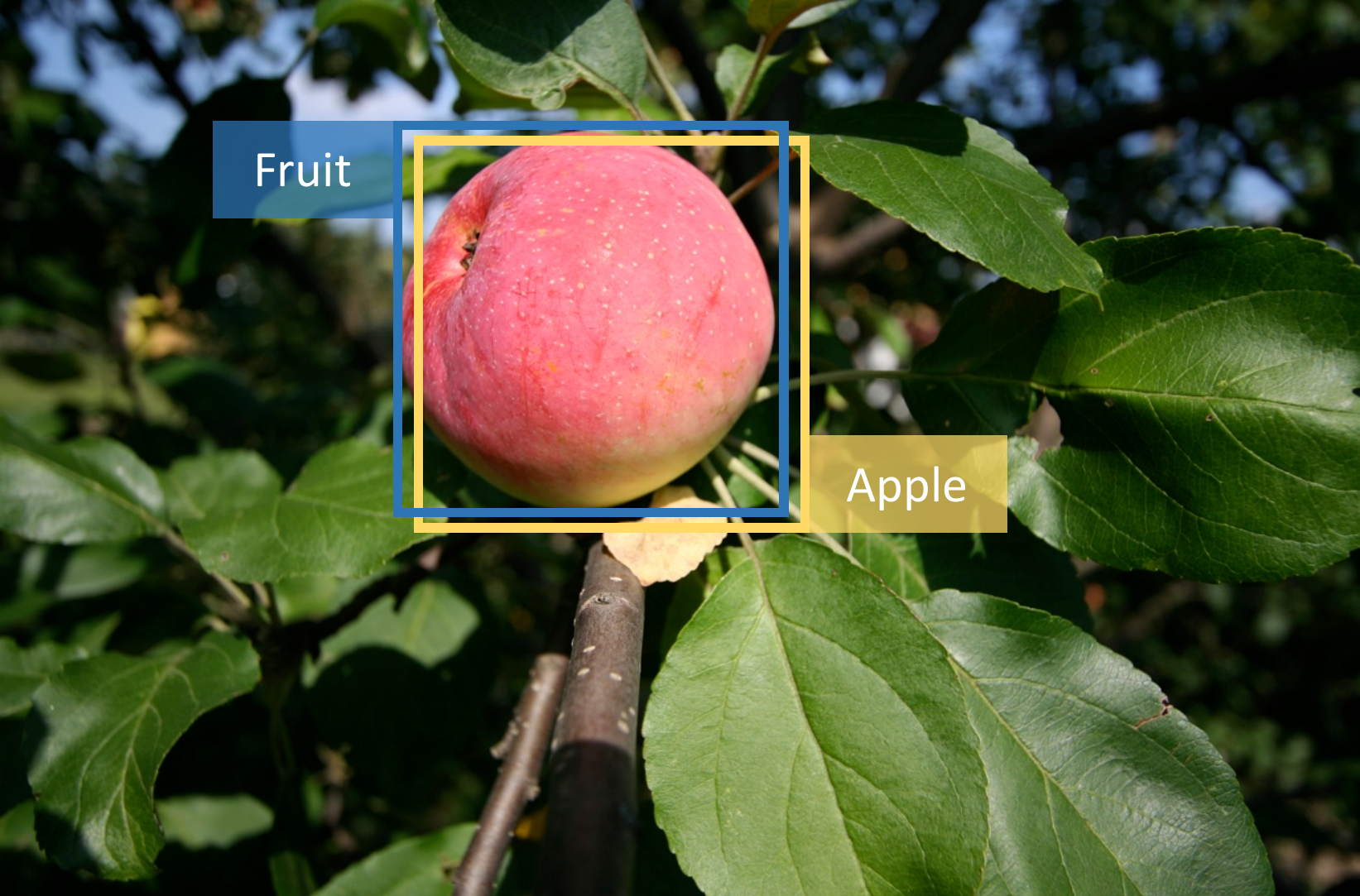}
\caption{}
\label{fig:problem_ex_hier}
\end{subfigure}
\begin{subfigure}[b]{0.247\textwidth}
\centering
\includegraphics[height=0.140\textheight]{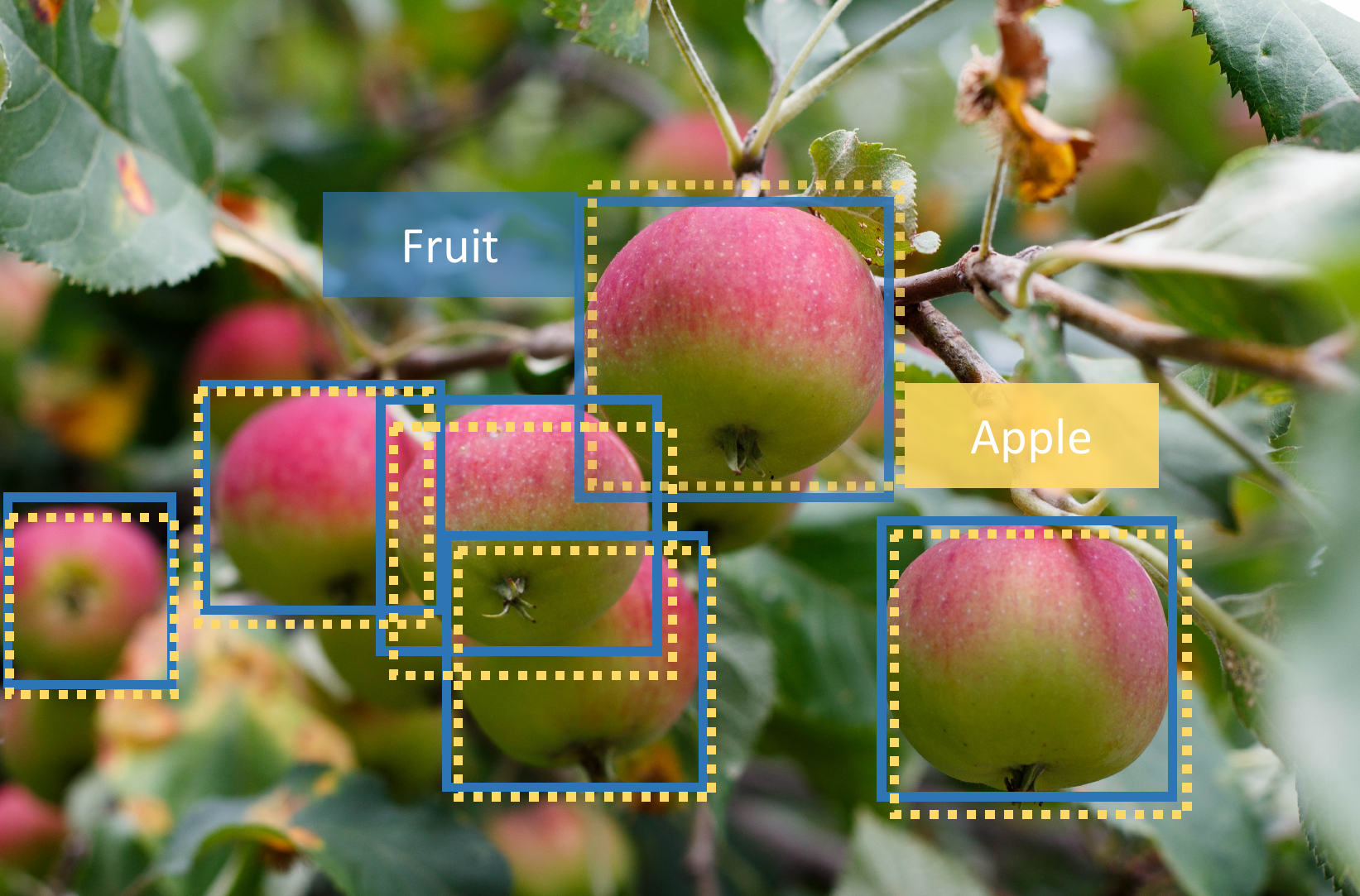}
\caption{}
\label{fig:problem_im_hier}
\end{subfigure}
\begin{subfigure}[b]{0.247\textwidth}
\centering
\includegraphics[height=0.140\textheight]{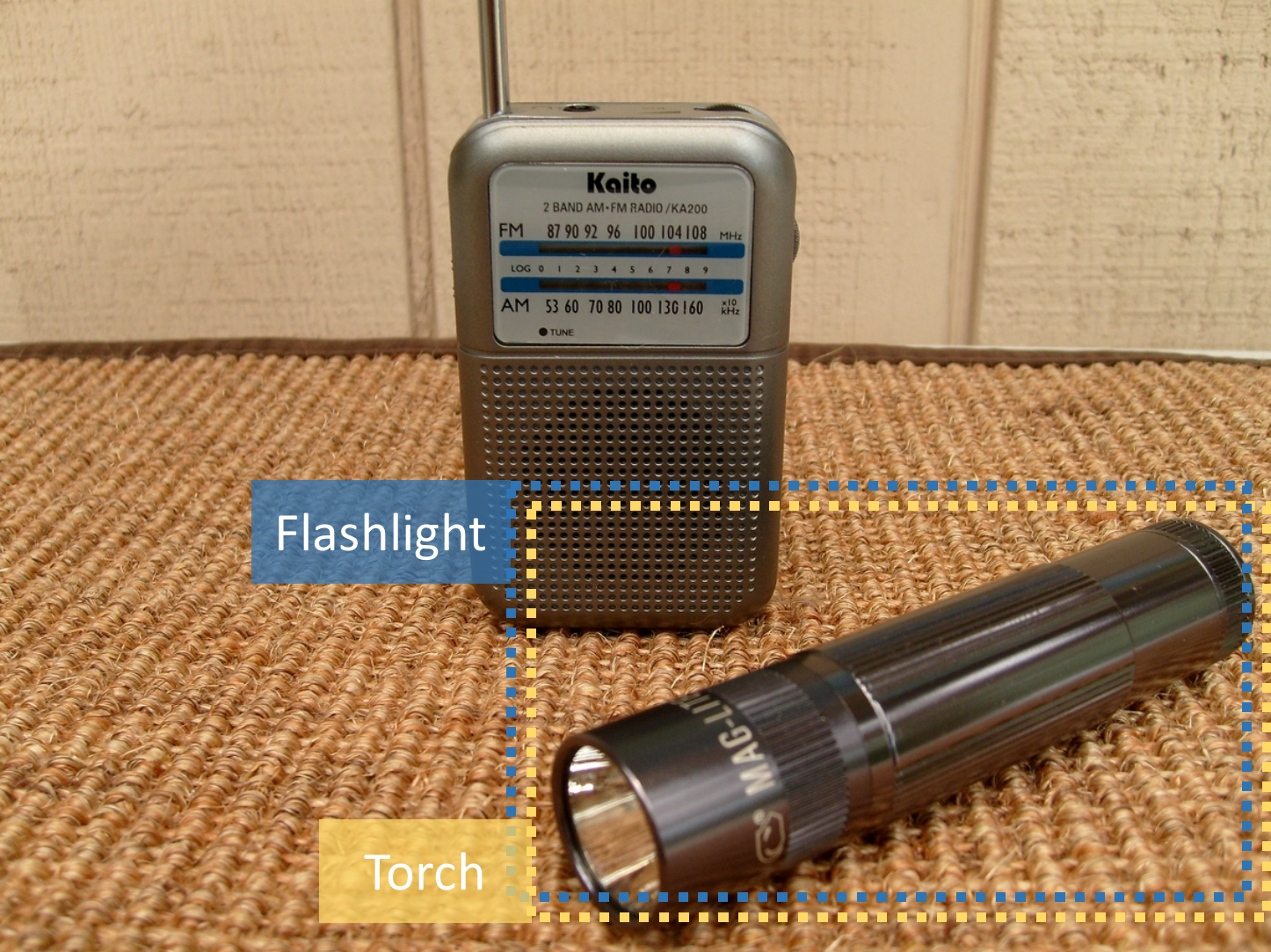}
\caption{}
\label{fig:problem_im_conf}
\end{subfigure}
\caption{Example of multi-label annotations in Open Images dataset.  (a)(b) are cases that objects are explicitly annotated with multiple labels. In (a), a car toy is labeled as car and toy at the same time. In (b), an apple is hierarchically labeled as apple and fruit. (c)(d) are cases that objects implicitly have multiple labels. In (c), apples are only labeled as fruit. In (d), flashlights are always randomly labeled as torch or flashlight.}
\label{fig:problems}
\end{figure*}

\begin{abstract}


Training with more data has always been the most stable and effective way of improving performance in deep learning era.
As the largest object detection dataset so far, Open Images brings great opportunities and challenges for object detection in general and sophisticated scenarios.  
However, owing to its semi-automatic collecting and labeling pipeline to deal with the huge data scale, Open Images dataset suffers from label-related problems that objects may explicitly or implicitly have multiple labels and the label distribution is extremely imbalanced. 
In this work, we quantitatively analyze these label problems and provide a simple but effective solution. We design a concurrent softmax to handle the multi-label problems in object detection and propose a soft-sampling methods with hybrid training scheduler to deal with the label imbalance. Overall, our method yields a dramatic improvement of 3.34 points, leading to the best single model with 60.90 mAP on the public object detection test set of Open Images. And our ensembling result achieves 67.17 mAP, which is 4.29 points higher than the best result of Open Images public test 2018.

\end{abstract}

\section{Introduction}






Data is playing a primary and decisive role in deep learning. With the advent of ImageNet dataset~\cite{deng2009imagenet}, deep neural network~\cite{krizhevsky2012imagenet} becomes well exploited for the first time, and an unimaginable number of works in deep learning sprung up. 
Some recent works~\cite{mahajan2018exploring, xie2019selftraining} also prove that larger quantities of data with labels of low quality(like hashtags) could surpass the state-of-the-art methods by a large margin. 
Throughout the history of deep learning, it could be easily learned that the development of an area is closely related to the data.

In the past years, great progresses have also been achieved in the field of object detection. Some generic object detection datasets with annotations of high quality like Pascal VOC~\cite{everingham2010pascal} and MS COCO~\cite{lin2014microsoft} greatly boost the development of object detection, giving birth to plenty of amazing methods~\cite{ren2015faster, redmon2016you, liu2016ssd, lin2017focal}. 
However, these datasets are quite small in today's view, and begin to limit the advancement of object detection area to some degree. 
Attempts are frequently made to focus on atomic problems on these datasets instead of exploring object detection in harder scenarios. 

%
Recently, Open Images dataset is published in terms of 1.7 million images with 12.4 million boxes annotated of 500 categories. This unseals the limits of data-hungry methods and may stimulate research to bring object detection to more general and sophisticated situations. However, accurately annotating data of such scale is labor intensive that manual labeling is almost infeasible. The annotating procedure of Open Images dataset is completed with strong assistance of deep learning that candidate labels are generated by models and verified by humans. 
This inevitably weakens the quality of labels because of the uncertainty of models and the knowledge limitation of human individuals, which leads to several major problems.

Objects in Open Image dataset may explicitly or implicitly have multiple labels, which differs from the traditional object detection.
The object classes in Open Images form a hierarchy that most objects may hold a leaf label and all the corresponding parent labels. However, due to the annotation quality, there are cases that objects are only labeled as parent classes and leaf classes are absent. Apart from hierarchical labels, objects in Open Images dataset may also hold several leaf classes like {\it car} and {\it toy}. Another annoying case is that objects of similar classes are frequently annotated as each other in both training and validation set, for example, {\it torch} and {\it flashlight}, as shown in~\ref{fig:problems}.

As images of Open Images dataset are collected from open source in the wild, the label distribution is extremely imbalanced that both very frequent and infrequent classes are included. Hence, methods for balancing label distribution are requested to be applied to train detectors better. Nevertheless, earlier methods like over-sampling tends to impose over-fitting on infrequent categories and fail to fully use the data of frequent categories.

In this work, we engage in solving these major problems in large-scale object detection. We design a concurrent softmax to deal with explicitly and implicitly multi-label problem. We propose a soft-balance method together with a hybrid training scheduler to mitigate the over-fitting on infrequent categories and better exploit data of frequent categories. Our methods yield a total gain of 3.34 points, leading to a 60.90 mAP single model result on the public test-challenge set of Open Images, which is 5.09 points higher than the single model result of the first place method ~\cite{akiba2018pfdet} on the public test-challenge set last year.
More importantly, our overall system achieves a 67.17 mAP, which is 4.29 points higher than their ensembled results.

\section{Related Works}
\paragraph{Object Detection.}
Generic object detection is a fundamental and challenging problem in computer vision, and plenty of works ~\cite{ren2015faster, redmon2016you, liu2016ssd, lin2017focal, dai2017deformable, lu2019grid, cai2018cascade, li2019scale, peng2019pod} in this area come out in recent years. Faster-RCNN~\cite{ren2015faster} first proposes an end-to-end two-stage framework for object detection and lays a strong foundation for successive methods. 
In~\cite{dai2017deformable}, deformable convolution is proposed to adaptively sample input features to deal with objects of various scales and shapes. 
~\cite{li2019scale, peng2019pod} utilize dilated convolutions to enlarge the effective receptive fields of detectors to better recognize objects of large scale.
~\cite{cai2018cascade, lu2019grid} focus on predicting boxes of higher quality to accommodate the COCO metric.

However, most works are still exploring in datasets like Pascal VOC and MS COCO, which are  small by modern standards.
Only a few works~\cite{niitani2019sampling, wu2018soft, akiba2018pfdet, gao2018solution} have been proposed to deal with large-scale object detection dataset like Open Images~\cite{kuznetsova2018open}. 
Wu {\it et al.}~\cite{wu2018soft} proposes a soft box-sampling method to cope with the partial label problem in large scale dataset. In~\cite{niitani2019sampling}, a part-aware sampling method is designed to capture the relations between entity and parts to help recognize part objects. 

\vspace{-0.5cm}
\paragraph{Multi-Label Recognition.}
There have been many amazing attempts~\cite{gong2013deep, wang2016cnn, zhang2018multilabel, li2014multi, tan2015learning, wu2017adaptive, chen2019multi} to solve the multi-label classification problem from different aspects. 
One simple and intuitive approach~\cite{boutell2004learning} is to transform the multi-label classification into multiple binary classification problems and fuse the results, but this neglects relationships between labels.
Some works~\cite{hu2016learning, li2016conditional, tan2015learning, wang2017multi} embed dependencies among labels with deep learning to improve the performance of multi-label recognition. In~\cite{li2014multi, tan2015learning, wu2017adaptive, chen2019multi}, graph structures are utilized to model the label dependencies. Gong {\it et al.}~\cite{gong2013deep} uses a ranking based learning strategy and reweights losses of different labels to achieve better accuracy. Wang {\it et al.}~\cite{wang2016cnn} proposes a CNN-RNN framework to embed labels into latent space to capture the correlation between them. 


\vspace{-0.5cm}
\paragraph{Imbalanced Label Distribution.}
There have been many efforts on handling long-tailed label distribution through data based resampling strategies~\cite{shen2016relay, gao2018solution, chawla2002smote, ouyang2016factors, zou2018unsupervised, mahajan2018exploring} or loss-based methods~\cite{cui2019class, huang2016learning,wang2017learning}.
In~\cite{shen2016relay, gao2018solution, ouyang2016factors}, class-aware sampling is applied that each mini-batch is filled as uniform as possible with respect to different classes.
~\cite{chawla2002smote, zou2018unsupervised} expand samples of minor classes through synthesizing new data. Dhruv {\it et al.}~\cite{mahajan2018exploring} computes a replication factor for each image based on the distribution of hashtags and duplicates images the prescribed number of times.

As for loss-based methods, loss weights are assigned to samples of different classes to match the imbalanced label distribution. In~\cite{huang2016learning,wang2017learning}, samples are re-weighted by inverse class frequency, while Yin {\it et al.}~\cite{cui2019class} calculates the effective number of each class to re-balance the loss. In OHEM~\cite{shrivastava2016training} and focal loss~\cite{lin2017focal}, difficulty of samples are evaluated in term of losses and hard samples are assigned higher loss weights. 


\begin{figure*}
\centering
\includegraphics[width=\textwidth]{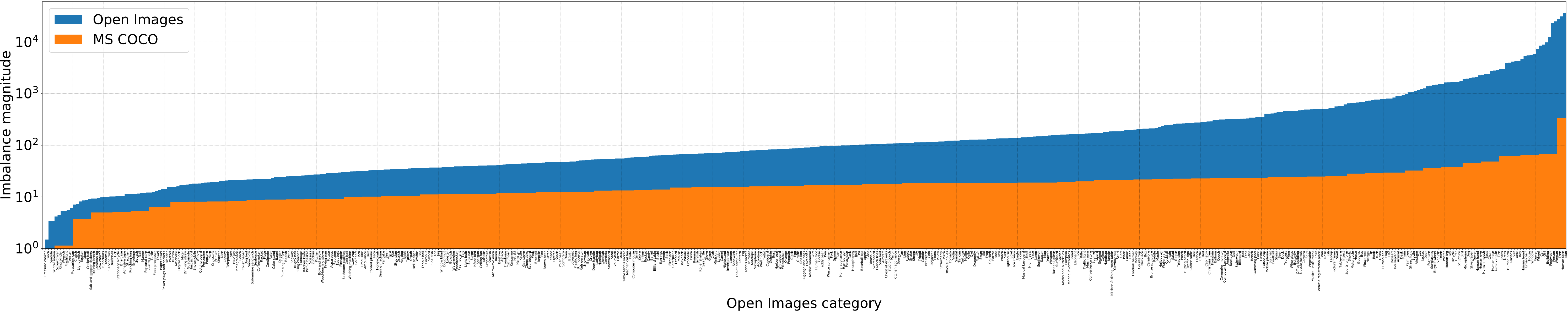}
\caption{
The imbalance magnitude of Open Images and MS COCO dataset.
{\it Imbalance magnitude} means the number of the images of the largest category divided by the smallest.
(best viewed on high-resolution display)
}
\label{fig:oic_imbalance}
\end{figure*}

\begin{figure}
\centering
\begin{subfigure}[b]{0.45\textwidth}
\centering
\includegraphics[width=\textwidth]{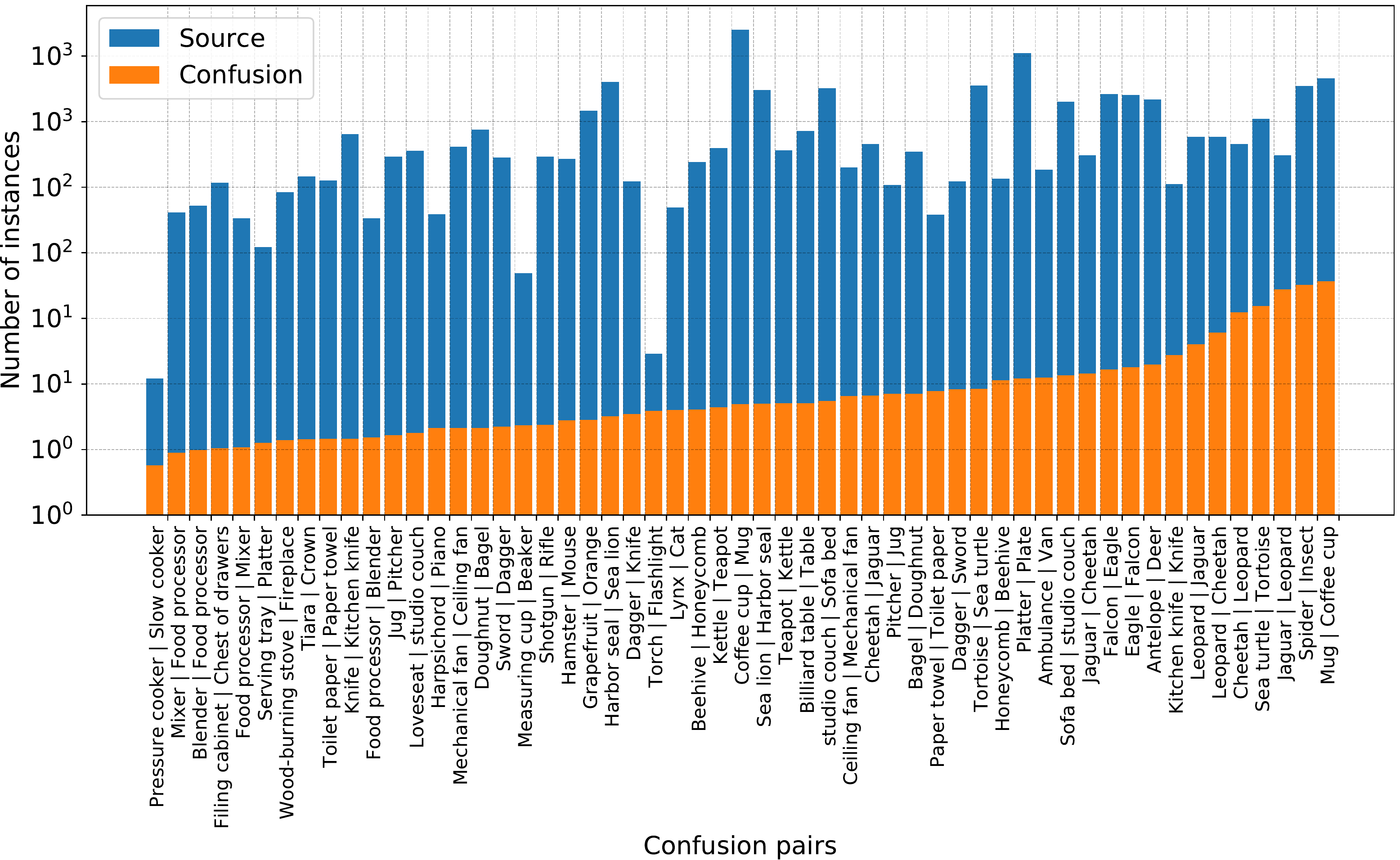}
\caption{We select the top-$55$ confused category pairs and show their concurrent rates.}
\label{fig:leaf_conf_dist}
\end{subfigure}
\centering

\begin{subfigure}[b]{0.45\textwidth}
\centering
\includegraphics[width=\textwidth]{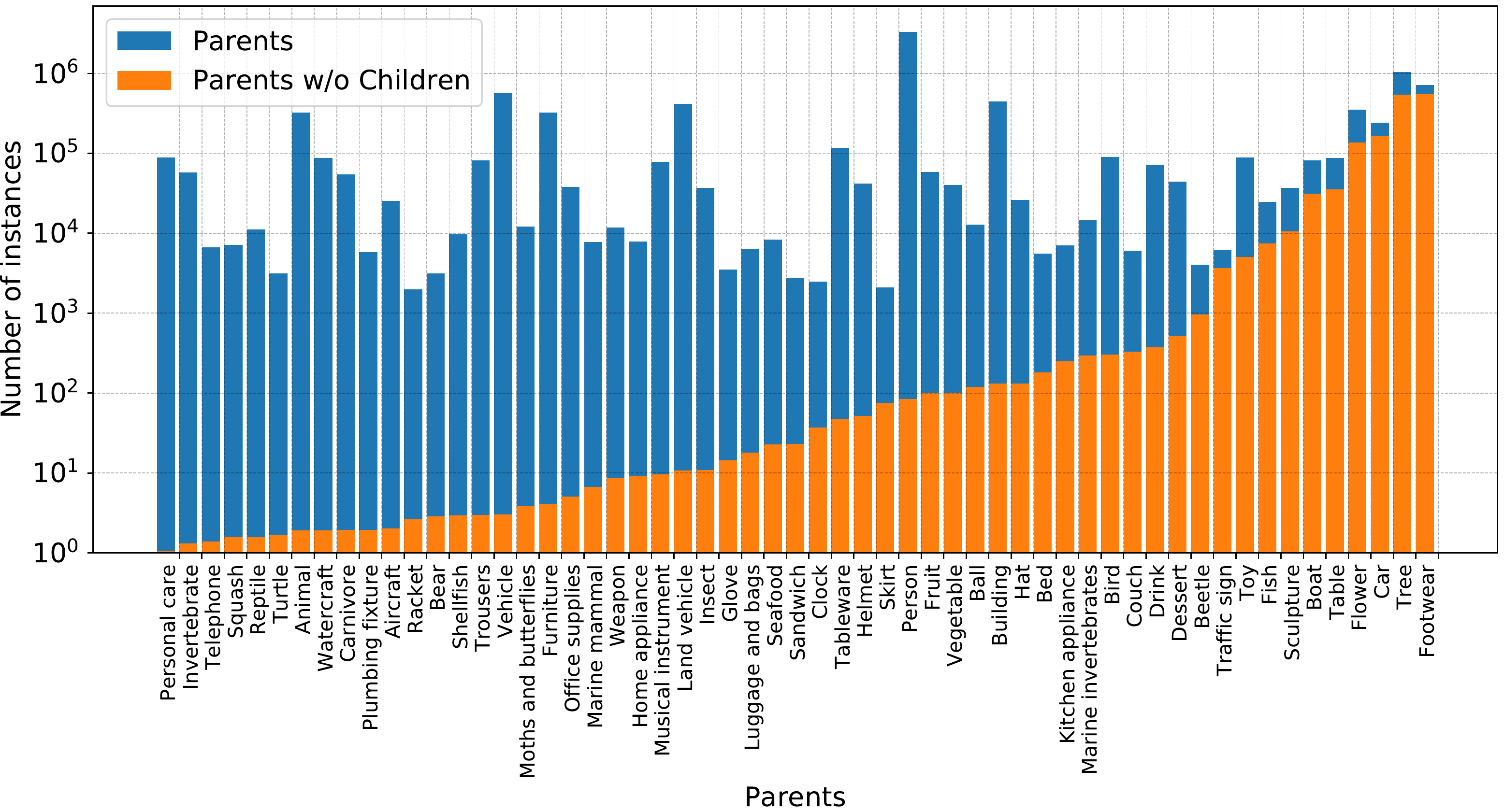}
\caption{We show the ratio of parent annotations without leaf label and total parent annotations.}
\label{fig:hier_conf_dist}
\end{subfigure}

\caption{
Implicit multi-label problem caused by confused categories and absence of leaf classes.}
\label{fig:show_dist}
\end{figure}

\section{Problem Setting}
\label{sec:problem_setting}
Open Images dataset is currently the largest released object detection dataset to the best of our knowledge. It contains 12 million annotated instances for 500 categories on 1.7 million images.
Considering its large size, it is unfeasible to manually annotate such huge number of images on 500 categories
Owing to its scale size and annotation styles, we argue that there are three major problems regarding this kind of dataset, apart from the missing annotation problem which has been discussed in ~\cite{wu2018soft, niitani2019sampling}.  

\noindent \textbf{Objects may explicitly have multiple labels.}
As objects in physical world always contain rich categorical attributes on different levels, 
the 500 categories in Open Images dataset form a class hierarchy of 5 levels, with 443 of the nodes as leaf classes and 57 of the nodes as parent classes. It is likely that some objects hold multiple labels including leaf labels and parent labels, like {\it apple} and {\it fruit} as shown in Figure~\ref{fig:problem_ex_hier}. Another case is that an object could easily have multiple leaf labels. For instance, a car-toy is labeled as {\it toy} and {\it car} at the same time as shown in Figure~\ref{fig:problem_ex_leaf}. This happens frequently in dataset and all leaf labels are requested to be predicted during evaluation. Different from previous single-label object detection, how to deal with multiple labels is one of the crucial factors for object detection in this dataset.

\noindent \textbf{Objects may implicitly have multiple labels.}
Other than the explicit multi-label problem, there is also the implicit multi-label problem caused by the limited and inconsistent knowledge of human annotators. There remain many pairs of leaf categories that are hard to distinguish, and labels of these pairs are mixed up randomly. We analyze the proportion that an object of a leaf class is labeled as another, and find that there are at least 115 pairs of severely confused categories(confusion ratio $\geq0.1$). We display the top-$55$ confused pairs in Figure~\ref{fig:leaf_conf_dist}, and find many categories heavily confused. For instance, nearly $65\%$ of the torches are labeled as {\it flashlight} and $50\%$ of the leopards are labeled as {\it cheetah}.

Besides, labels of leaf and parent classes are always not complete so that a large amount of objects are only annotated with parent labels without leaf label. As shown in Figure~\ref{fig:problem_im_hier}, an apple is sometimes labeled as {\it apple} and sometimes labeled as {\it fruit} without the leaf annotation. We demonstrate the ratio of leaf annotations and parent annotations lacking leaf annotations in Figure~\ref{fig:hier_conf_dist}. This implicit co-existence phenomenon also happens frequently and needs to be taken into consideration, otherwise the detectors may learn the false signals.

\noindent \textbf{Imbalanced label distribution.}
\label{sec:problem-imbalance}
To build such a huge dataset, images are collected from open source in the wild.
As on can expect, Open Images dataset suffers from extremely imbalanced label distribution that both infrequent and very frequent categories are included. As shown in Figure~\ref{fig:oic_imbalance}, the most frequent category owns nearly $30$k times the training images of the most infrequent category.
Naive re-balance strategy, such as widely used class-aware sampling~\cite{gao2018solution} which uniformly samples training images of different categories could not cope with such extreme imbalance, and may lead to two consequences: 

\noindent 1) 
For frequent categories,
they are not trained sufficiently, for the reason that most of their training samples have never been seen and are wasted. 

\noindent 2) 
For infrequent categories,
the excessive over-sampling on them may cause severe over-fitting and degrade the generalizability of recognition on these classes. 

Once adopting the class-aware sampling, category like {\it person} is extremely undersampled that $99.13\%$ of the instances are neglected, while category like {\it pressure cooker} is 
immensely
oversampled that each instance is seen for $252$ more times averagely within an epoch.

\section{Methodology}
\label{sec:methodology}
In this part, we explore methods to deal with the label related problems in large scale object detection.
First, we design a {\it concurrent softmax} to handle both the explicit and implicit multi-label issues jointly. Second, we propose a {\it soft-balance sampling} together with a {\it hybrid training scheduler} to deal with the extremely imbalanced label distribution.

\subsection{Multi-Label Object Detection}
As one of the most widely used loss function in deep learning, the form of softmax loss about a bounding box $b$ is presented as follows:
\begin{equation}
L_{cls}(b) = -\sum\limits_{i=1}\limits^{C}y_i\log(\sigma_{i}),\;\;\; with \; \sigma_{i} = \frac{e^{z_i}}{\sum_{j=1}^{C}{e^{z_j}}}, 
\end{equation}
where $z_i$ denotes the response of the $i$-th class, $y_i$ denotes the label and $C$ means the number of categories. It behaves well in single label recognition where $\sum{y_i}=1$.
However, things are different when it comes to multi-label recognition.

In the conventional object detection training scheme, each bounding box is assigned only one label during training ignoring other ground-truth labels.
If we force to assign all the $m$($m\geq1$) ground-truth labels that belongs to $K = \{k \mid k \in C , y_k = 1\}$ to bounding box during training, scores of multiple labels would restrain each other. When computing the gradient of each ground-truth label, it looks like below:
\begin{equation}
\frac{\partial L_{cls}}{\partial z_i} = 
    \begin{cases}
    m\sigma_i - 1, & \text{if} \; i \in K; \\
    m\sigma_i,     & \text{if} \; i \notin K.
    \end{cases}
\end{equation}
When $m\sigma_i > 1$ for $i \in K$, $z_i$ is optimized to become lower even if $i$ is one of the ground-truth labels, which is the wrong optimization direction.

 

\subsubsection{Concurrent Softmax}
The {\it concurrent softmax} is designed to help solve the problem of recognizing objects with multiple labels in object detection. During training, the concurrent softmax loss of a predicted box $b$ is presented as follows:
\begin{equation}
\begin{split}
L_{cls}^*(b) &= -\sum\limits_{i=1}\limits^{C}y_i\log{\sigma_i^*}, \\
with \; \sigma_i^* &= \frac{e^{z_i}}{\sum_{j=1}^{C}{(1-y_{j})(1 - r_{ij})e^{z_j} + e^{z_i}}},   
\end{split}
\end{equation}
where $y_{i}$ denotes the label of class $i$ regarding the box $b$, and $r_{ij}$ denotes the concurrent rate of class $i$ to class $j$. And output of concurrent softmax during training is defined as:
\begin{equation}
\frac{\partial L^*_{cls}}{\partial z_i} = 
    \begin{cases}
    \sigma_i^* - 1, & \text{if} \; i \in K; \\
    \sum\limits_{j\in K}{(1-r_{ij})\sigma_i^*},     & \text{if} \; i \notin K.
    \end{cases}
\end{equation}

Unlike in softmax that responses of the ground-truth categories are suppressing all the others, we remove the suppression effects between explicitly coexisting categories in concurrent softmax. For instance, a bounding box is assigned multiple ground-truth labels $K = \{k \mid k \in C , y_k = 1\}$ during training. When computing the score of class $i\in K$, influences of all the other ground-truth classes $j \in K \setminus \{i\}$ are neglected because of the  $(1-y_{j})$ term, and the score of each correct class is boosted. This avoids the unnecessary large losses due to the multi-label problem, and the gradients could focus on more valuable knowledge.

Apart from the explicit co-existence cases, there are still implicit concurrent relationships remain to be settled. 
We define a concurrent rate $r_{ij}$ as the probability that an object of class $i$ is labeled as class $j$. The $r_{ij}$ is calculated based on the class annotations of training set and Figure~\ref{fig:leaf_conf_dist} shows the concurrent rates of confusion pairs. For hierarchical relationships, $r_{ij}$ is set 0 when $i$ is leaf node with $j$ as its parent, and vice versa. With the $(1-r_{ij})$ term, suppression effects between confusing pairs are weakened. 




The influence of multi-label object detection is also prominent during inference. Different from the conventional multi-label recognition tasks, the evaluation metric of object detection is mean average precision(mAP). For each category, detection results of all images are firstly collected and ranked by scores to form a precision-recall curve, and the average area of precision-recall curve is defined as the mAP. In this way, the absolute value of box score matters, because it may influence the rank of predicted box over the entire dataset.
Thus we also apply the concurrent softmax during inference, and  present it as follows:
\begin{equation}
 \sigma_i^{\dagger} = \frac{e^{z_i}}{\sum_{j=1}^{C}{(1 - r_{ij})e^{z_j}}},
\end{equation}
where we abandon the $(1-y_j)$ term and keep the concurrent rate term.
Scores of categories in a hierarchy and scores of similar categories would not suppress each other, and are boosted effectively, which is desirable in object detection task.

\subsubsection{Compared with BCE loss}
\label{sec:other_losses}
BCE is always a popular solution to mutl-label recognition, but it does not work well on multi-label detection task. We argue that sigmoid function fails to normalize scores and declines the suppression effect between categories which is desired when evaluated with mAP metric. We have tried BCE loss and focal loss, but it turns out that they yield much worse result even than the original softmax cross-entropy.

\subsection{Soft-balance Sampling with Hybrid Training}
As detailedly illuminated in \ref{sec:problem-imbalance}, Open Images dataset suffers from severely imbalanced label distribution.
We denote by $C$ the number of categories, $N$ the number of total training images,
and $n_i$ the number of images containing objects of the $i$-th class.
Conventionally, images are sampled in sequence without replacement for training in each epoch,
and the original probability $P_o$ of class $i$ being sampled is denoted as $P_{o}(i) = \frac{n_i}{N}$, 
which may greatly degrade the recognition capability of model for infrequent classes.
A widely used technique, \ie, class-aware sampling mentioned in~\cite{shen2016relay, gao2018solution, ouyang2016factors} is a naive solution to handle the class imbalance problem,
in which categories are sampled uniformly in each batch.
The class-aware sampling probability $P_{a}$ of class $i$ becomes $P_{a}(i) = \frac{1}{C}$.
Yet this may cause heavy over-fitting on infrequent categories and insufficient training on frequent categories as aforementioned.

To alleviate the problems above, we firstly adjust the sampling probability based on number of samples, which we call {\it soft-balance sampling}.
We first define the $P_{n}(i) = \frac{n_i}{\sum_{i=j}^{C}{n_j}}$ as the approximation of non-balance sampling probability $P_{o}(i)$ for convenience.
Then the sampling probability of class-aware balance can be reformulated as:
\begin{equation}
\label{eq:p_h}
\begin{aligned}
P_{a}(i) & = \frac{1}{C} = \frac{1}{C P_{n}(i)} P_{n}(i) = \alpha P_{n}(i),
\end{aligned}
\end{equation}
where the $\alpha = \frac{1}{C P_{n}(i)}$ can be regarded as a balance factor that is inversely proportional to the number of categories and the original sampling probability.

To reconcile the frequent and infrequent categories, we introduce soft-balance sampling by adjusting the balance factor with a new hyper-parameter $\lambda$:
\begin{equation}
\label{eq:p_s}
\begin{aligned}
P_{s}(i) &= \alpha^{\lambda} P_{n}(i) \\
& = P_{a}(i)^{\lambda} P_{n}(i)^{(1-\lambda)}.
\end{aligned}
\end{equation}
Note that $\lambda = 0$ corresponds to non-balance sampling and $\lambda = 1$ corresponds to class-aware balance.
The normalized probability is:
\begin{equation}
\label{eq:p_s_n}
\begin{aligned}
P_{s}^*(i) = \frac{P_{s}(i)}{\sum_{j=1}^{C}{P_{s}(j)}}
\end{aligned}.
\end{equation}
This sampling strategy guarantees more sufficient training on dominate categories and decreases the excessive sampling frequency of infrequent categories.

Even with the soft-balance method, there are still many samples of the frequent categories that are not sampled.
Thus we propose a hybrid training scheduler to further mitigate this problem.
We firstly train detector using the conventional strategy,
which is sampling training images in sequence without replacement, and the equivalent sampling probability is $P_o$.
Then we finetune the model with soft-balance strategy to cover categories with very few samples.
This hybrid training schema exploits the effectiveness of pretrained model for object detection task from Open Images itself rather than ImageNet.
It ensures that all the images have been seen during training, and endows the model with a better generalization ability.



\section{Experiments}
\label{sec:experiments}

\subsection{Dataset}


To analyze the proposed concurrent softmax loss and soft-balance with hybrid training, we conduct experiment on Open Images challenge 2019 dataset.
As an object detection dataset in the wild,
it contains 1.7 million images with 12.4 million boxes of 500 categories in its challenge split.
The scale of training images is $15$ times of the MS COCO~\cite{lin2014microsoft} and $3$ times of the second largest object detection dataset Object365~\cite{shao2019objects365}.

Considering the huge size of Open Images dataset,
we split a mini Open Images dataset for our ablation study.
The mini Open Images dataset contains 115K training images and 5K validation images named as {\it \textcolor{blue}{mini-train}} and {\it \textcolor{blue}{mini-val}}.
All the images are sampled from Open Images challenge 2019 dataset with the ratio of each category unchanged.
Final results on {\it full-val} and {\it public test-challenge} in Open Images challenge 2019 dataset are also reported.
We follow the metric used in Open Images challenge which is a variant mAP at IoU 0.5, as all false positives not existing in image-level labels\footnote{In Open Images dataset, image-level labels consist of verified-exist labels and verified-not-exist labels. The unverified categories are ignored} are ignored.


\subsection{Implementation Details}
We train our detector with ResNet-50 backbone armed with FPN.
For the network configuration, we follow the setting mentioned in Detectron.
We use SGD with momentum 0.9 and weight decay 0.0001 to optimize parameters.
The initial learning rate is set to 0.00125 $\times $ batch size, and then decreased by a factor of 10 at 4 and 6 epoch for $1 \times$ schedule which has total 7 epochs.
The input images are scaled so that the length of the shorter edge is 800 and the longer side is limited to 1333.
Horizontal flipping is used as data augmentation and sync-BN is adopted to speed up the convergence.

\subsection{Concurrent Softmax}
We explore the influence of concurrent softmax in training and testing stage respectively in this ablation study. 
All models are trained with {\it mini-train} and evaluated on {\it mini-val}.

\noindent \textbf{The impacts of concurrent softmax during training.} 
Table~\ref{tab:loss-train} shows the results of the proposed concurrent softmax compared with the vanilla softmax and other existing methods during training stage. Concurrent softmax could outperform softmax by 1.13 points with class-aware sampling and 0.98 points with non-balance sampling. It is also found that sigmoid with BCE and focal loss behaves poorly in this case. We guess that they are incompatible with the mAP metric in object detection as mentioned in~\ref{sec:other_losses}. Our method also outperforms dist-CE loss~\cite{mahajan2018exploring} and Co-BCE loss~\cite{akiba2018pfdet}.

\begin{table}
\begin{center}
\caption{
The comparison of different loss functions method. Models are trained in {\it \textcolor{blue}{mini-train}} and evaluated on {\it \textcolor{blue}{mini-val}}.}

\label{tab:loss-train}
\begin{tabular}{l|c|c}
\hline
Loss Type & Balance & mAP    \\
\hline
\hline
 Focal Loss~\cite{lin2017focal} & \cmark & 50.18 \\
 BCE Loss & \cmark &  54.29 \\
 Co-BCE Loss~\cite{akiba2018pfdet} & \cmark & 55.74 \\
 dist-CE Loss~\cite{mahajan2018exploring} & \cmark  & 55.90 \\
\hline
Softmax Loss &        & 38.16 \\ 
Concurrent Softmax Loss &         & \textbf{39.14} \\
Softmax Loss &  \cmark & 55.45 \\ 
Concurrent Softmax Loss &  \cmark & \textbf{56.58} \\ 
\hline
\end{tabular}
\end{center}
\end{table}

\noindent \textbf{The impacts of concurrent softmax during testing.} 
We also show results to demonstrate the effectiveness of concurrent softmax in testing stage in Table~\ref{tab:loss-inference}. Solely applying concurrent softmax brings 0.36 mAP improvement during inference, while applying it in both training and testing stage yields 1.50 points improvement totally. This also proves the fact that suppression effects between leaf and parent categories or confusing categories are harming the performance of object detection in Open Images.

\begin{table}
\begin{center}
\caption{
The effectiveness of concurrent softmax during testing. Models are trained in {\it \textcolor{blue}{mini-train}} and evaluated on {\it \textcolor{blue}{mini-val}}.}

\label{tab:loss-inference}
\begin{tabular}{cc|c}
\hline
Train Method & Test Method & mAP    \\
\hline
\hline
Softmax & Softmax                       & 55.45 \\ 
Softmax & Concurrent Softmax            & \textbf{55.77} \\
Concurrent Softmax & Softmax            & 56.58 \\ 
Concurrent Softmax & Concurrent Softmax & \textbf{56.95} \\ 

\hline
\end{tabular}
\end{center}

\end{table}

\subsection{Soft-balance Sampling}

\begin{table}
\begin{center}
\caption{
The comparison of different sampling methods. Models are trained in {\it \textcolor{blue}{mini-train}} and evaluated on {\it \textcolor{blue}{mini-val}}.}
\label{tab:power}
\begin{tabular}{l|c|c}
\hline
Methods & $\lambda$     & mAP    \\
\hline
\hline
Non-balance &-& 38.16 \\
Class-aware Sampling~\cite{gao2018solution} &-& 55.45\\
Effective Number~\cite{cui2019class} &-& 45.72 \\
\hline
\multirow{5}{6em}{Soft-balance} & 0.3 & 50.69 \\
& 0.5 & 56.19 \\
& 0.7 &\textbf{57.04} \\
& 1.0 & 55.45 \\
& 1.5 & 52.41 \\
\hline
\end{tabular}
\end{center}
\end{table}

\begin{figure}[t]
\begin{center}
\includegraphics[width=0.99\linewidth]{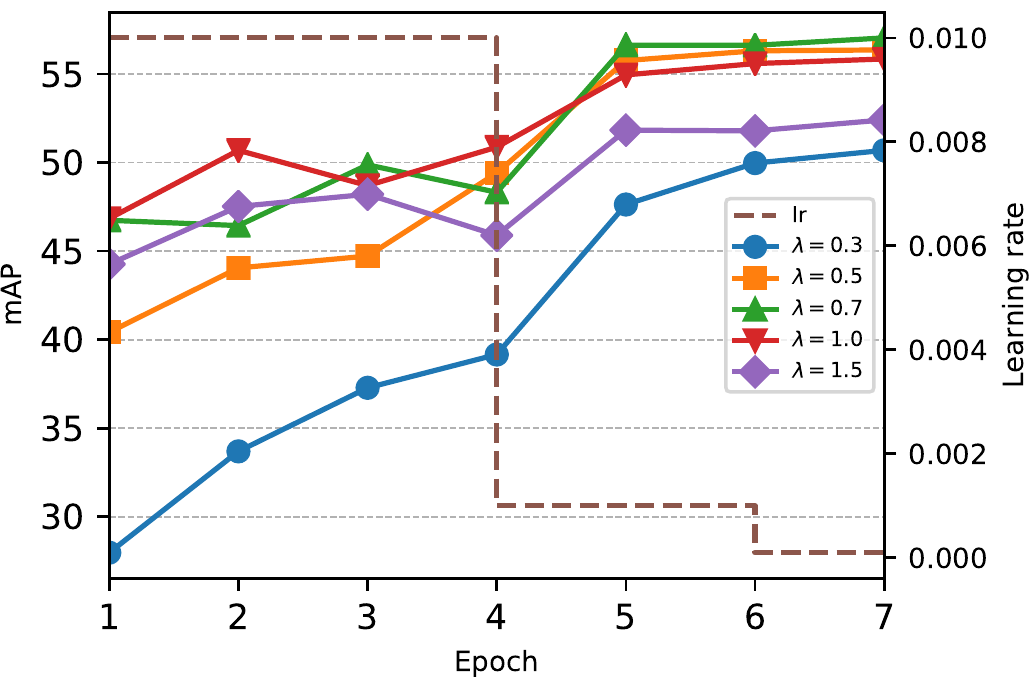}
\end{center}
\caption{Training curves of the proposed soft-balance sampling. Soft-balance with $\lambda=0.7$ achieves the best performance.}
\label{fig:power}
\end{figure}





\begin{figure*}
\centering
\begin{subfigure}[b]{0.95\linewidth}
\begin{minipage}{1\linewidth}
\centering
\begin{subfigure}[b]{0.49\linewidth}
\centering
\includegraphics[width=\textwidth]{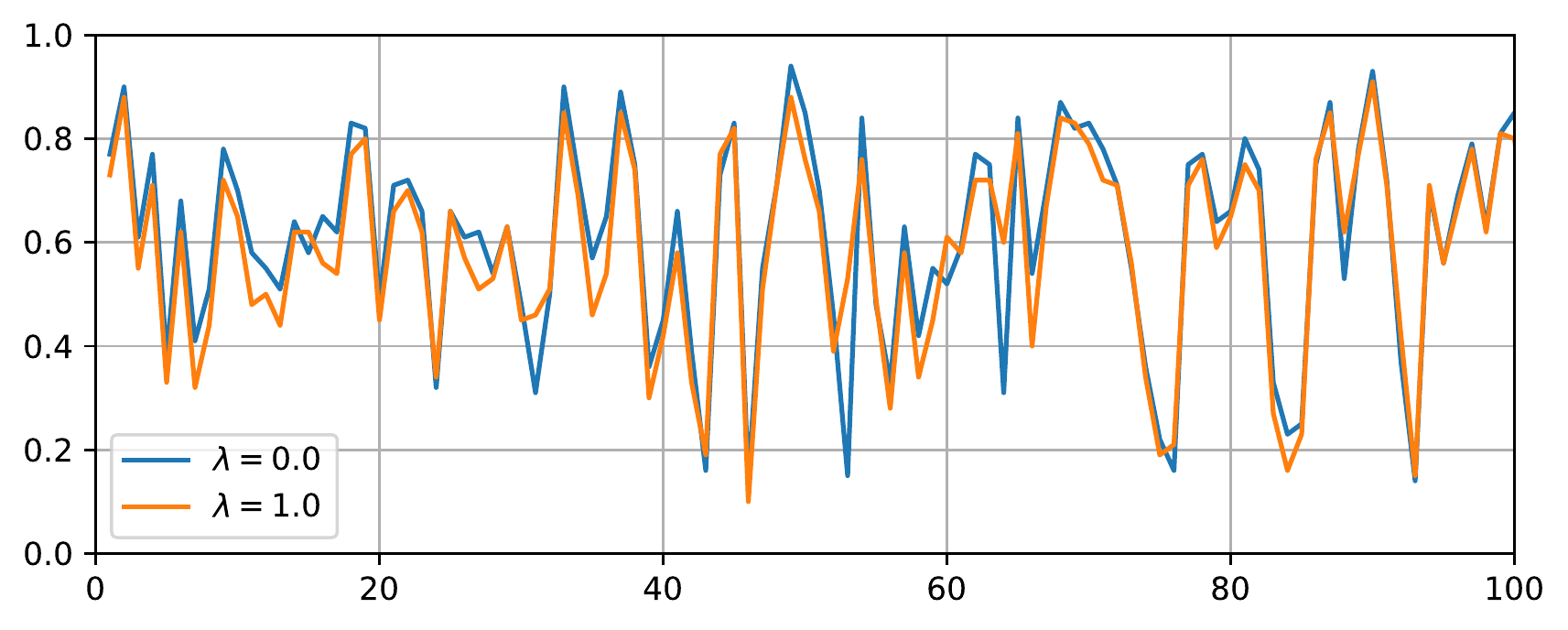}
\end{subfigure}
\begin{subfigure}[b]{0.48\linewidth}
\centering
\includegraphics[width=\linewidth]{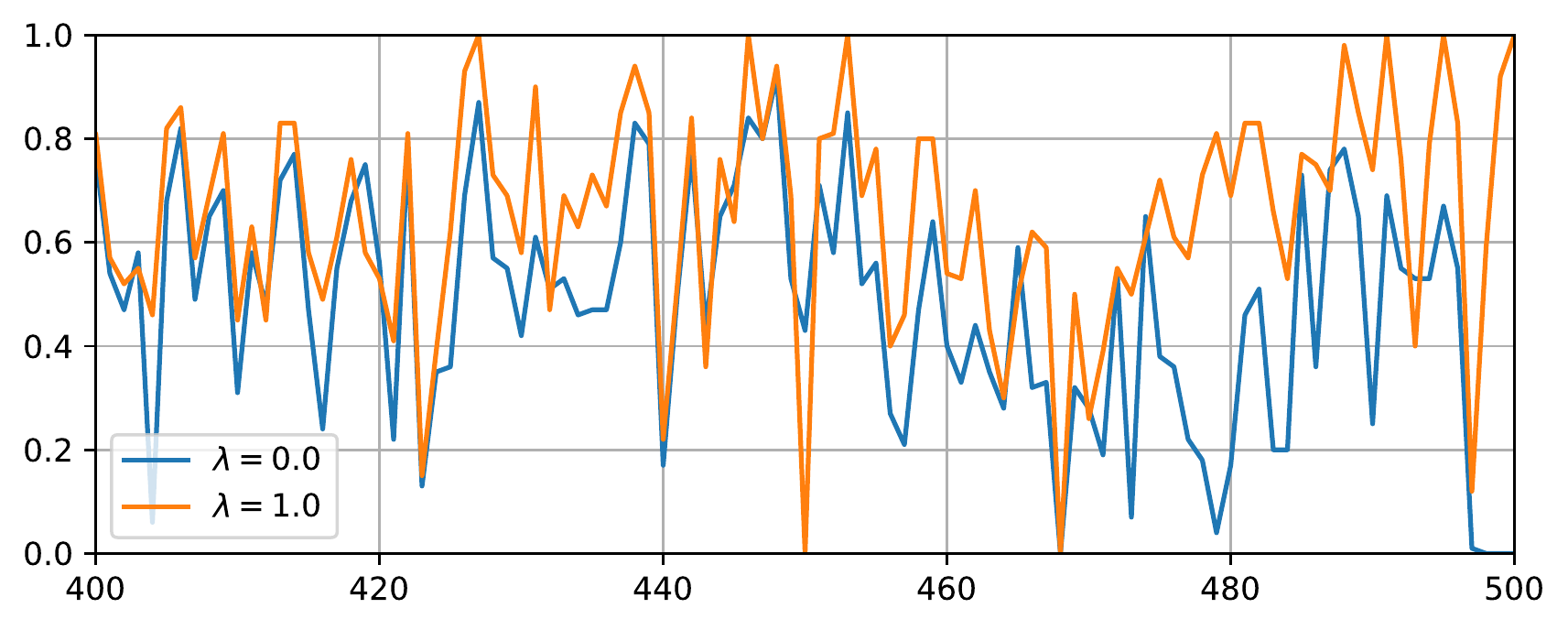}
\end{subfigure}
\caption{Non-balance (blue) versus Class-aware Sampling (orange) sorted by the number of images for most frequent 100 categories (left) and most infrequent 100 categories (right).
}
\label{fig:nb_hn_img}
\end{minipage}
\end{subfigure}

\begin{subfigure}[b]{0.95\linewidth}
\begin{minipage}{1\linewidth}
\centering
\begin{subfigure}[b]{0.49\linewidth}
\centering
\includegraphics[width=\textwidth]{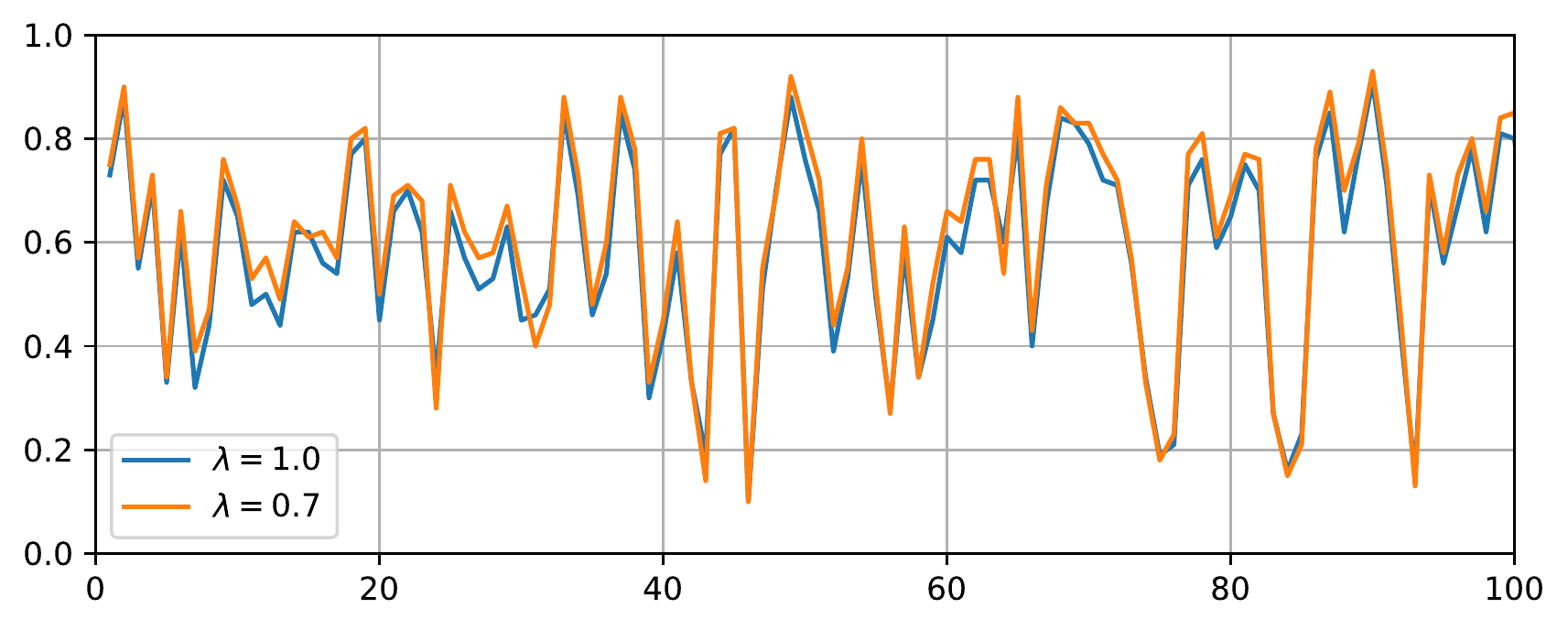}
\end{subfigure}
\begin{subfigure}[b]{0.48\linewidth}
\centering
\includegraphics[width=\linewidth]{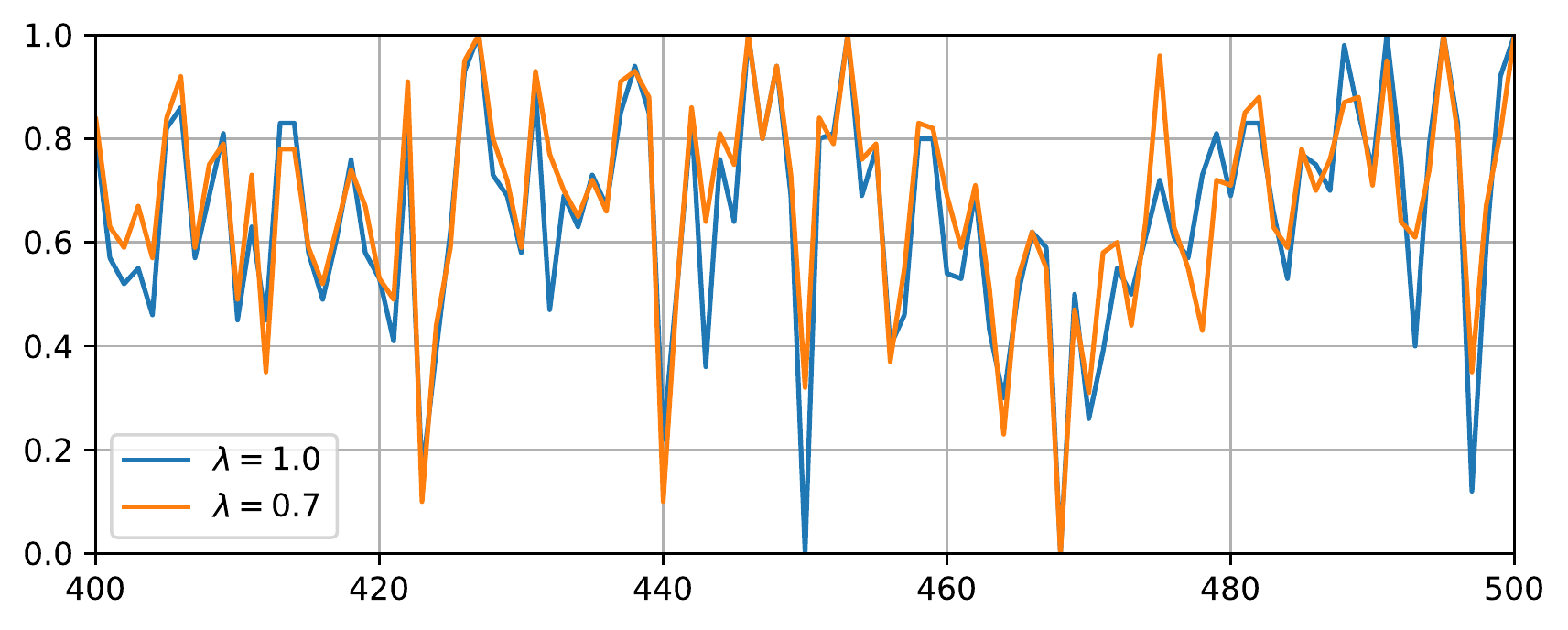}
\end{subfigure}
\caption{Class-aware Sampling (blue) versus Soft-balance with $\lambda=0.7$ (orange) sorted by the number of images for most frequent 100 categories (left) and most infrequent 100 categories (right).
}
\label{fig:hb_sb_img}
\end{minipage}
\end{subfigure}

\caption{
The comparison of sampling strategy among categories.
(Best viewed on high-resolution display)
}
\label{fig:balance_on_categories}
\end{figure*}

\noindent \textbf{Results.}
Table~\ref{tab:power} presents the results of the proposed soft-balance sampling and other balance method.
As Open Images is a long-tailed dataset, many categories have few samples, so that non-balance training only achieves 38.16 mAP.
Class-aware sampling simply samples all categories data uniformly at random, it remedy the data imbalance problem to a great extent and boost the performance to 55.45.
The effective number~\cite{cui2019class} is used to re-weight the classification loss with the purpose of harmonizing the gradient contribution from different categories.
Comparing to the non-balance method, the effective number improves the results by 7.56 points, but is worse than the class-aware sampling.
We argue that it is because the balance strategy applied on data level is more efficient than that of loss level.
Soft-balance with hyper-parameter $\lambda$ allows us to transfer from non-balance ($\lambda=0$) to class-aware sampling ($\lambda=1$).
Thus, we can find a point at which sufficient infrequent categories data could be sampled to train the model and the over-fitting problem does not happen yet.
The soft-balance with $\lambda=0.7$ outperforms the class-aware sampling by 1.59 points.

\noindent \textbf{The impacts of soft-balance during training.}
To investigate why soft-balance is better, we show the training curves of soft-balance with different $\lambda$ in Figure~\ref{fig:power}.
We can learn that the small $\lambda=0.3$ is hard to achieve a good performance due to the data imbalance problem.
But too large $\lambda=1.5$ also fails on accomplishing the best performance comparing with the relatively smaller $\lambda$ setting.
Note that the mAP of $\lambda=1.5$ is much higher than that of $\lambda=0.5$ in the first learning rate stage (before epoch 4),
but this situation reverses in subsequent train progress.
This comparison proves that $\lambda=1.5$ provides more sufficient rare categories data to train the model and achieve better performance in the beginning, however, it run into a severe over-fitting in the convergence stage.
The results of $\lambda=1.0$ and $\lambda=0.7$ validate this rules again.

\noindent \textbf{The impacts of soft-balance among categories.}
We further study the performance of $\lambda=0.0$, $\lambda=0.7$, and $\lambda=1.0$ among categories in Figure~\ref{fig:balance_on_categories},
in which the challenge validation results of 500 categories are arranged from large to few by their number of images.
As shown in Figure~\ref{fig:nb_hn_img}, the $\lambda=1.0$ (orange line) outperforms the $\lambda=0.0$ (blue line) on the later half categories which have few image samples.
Although $\lambda=1.0$ solves the data insufficiency of infrequent categories,
it under-samples the frequent categories and causes the performance dropping on the former half categories.
Figure~\ref{fig:hb_sb_img} shows that $\lambda=0.7$ (orange line) alleviates the excessive under-sampling of the major categories comparing to $\lambda=1.0$ (blue line).
On the other hand, it mitigates the over-fitting problem of infrequent categories.
Therefore, the performance of $\lambda=0.7$ is almost always better than $\lambda=1.0$ on the full category space.

\subsection{Hybrid Training Scheduler}

\begin{table}
\begin{center}
\caption{
The effect of training scheduler.
The $\lambda$ of the soft-balance is set to 0.7.
{\it Non-balance I14} denotes the model of epoch 14 trained with non-balance strategy from ImageNet pretrain.
{\it Non-balance S20} denotes the model of epoch 20 trained with non-balance strategy from scratch.
Soft-balance$^*$ means that concurrent softmax is adopted in both training and testing stage. Models are trained on {\it \textcolor{blue}{full-train}} and evaluated on {\it \textcolor{blue}{full-val}}.
}
\label{tab:hybrid}
\begin{tabular}{l|c|c|c}
\hline
Method & Pretrain & Epochs & mAP    \\
\hline
\hline
\multirow{5}{6em}{Non-balance}
& ImageNet & 7  & 56.06 \\
& ImageNet & 11 & 59.12 \\
& ImageNet & 14 & 59.85 \\
& ImageNet & 16 & 59.95 \\
& Scratch  & 20 & 60.70 \\
\hline
\multirow{4}{6em}{Class-aware Sampling}
& ImageNet & 7  & 64.68 \\
& ImageNet & 14 & 62.85 \\
& Non-balance I14 & 14+7 & 65.60 \\
& Non-balance S20 & 20+7 & 65.92 \\
\hline
\multirow{1}{6em}{Soft-balance}
& Non-balance S20 & 20+7 & \textbf{67.09} \\
\multirow{1}{6em}{Soft-balance$^*$}
& Non-balance S20 & 20+7 & \textbf{68.23} \\
\hline
\end{tabular}
\end{center}
\end{table}

Table~\ref{tab:hybrid} summarizes the results of ResNeXt152 on Open Images Challenge dataset trained with different training scheduler.
For the non-balance setting, the more epochs the model trained, the better performance the model achieves.
And training a model from scratch yields better results than finetuning from ImageNet pretrained model.
These observations match similar conclusion in~\cite{he2018rethinking}.

While class-aware sampling significantly boosts the performance by 8.62 points using the ImageNet pretraining in 7 epochs setting,
it suffers from over-fitting problem, as the mAP of model trained with 14 epochs is lower than that of 7 epochs.
And frequent categories are still intensely under-sampled even applying the balance sampling.
With hybrid training, class-aware sampling can achieve better performance in both non-balance ImageNet pretraining and non-balance scratch pretraining setting.
Note that these improvements are not caused by more training epochs,
because longer training schedule will decreases the performance if with only ImageNet pretraining.
By further using soft-balance strategy, the hybrid training with non-balance scratch is improved from 65.92 to 67.09 mAP.

\subsection{Extension Results on Test-challenge Set}

\begin{table}
\begin{center}
\caption{
Results with bells and whistles on Open Images public test-challenge set.
}
\label{tab:challenge}
\begin{tabular}{l|c|c}
\hline
Methods & Ensemble & Public Test  \\
\hline
\hline
2018 1st~\cite{akiba2018pfdet} & & 55.81 \\
Ours                        &  &\textbf{60.90} \\
\hline
2018 1st~\cite{akiba2018pfdet} & \cmark & 62.88 \\
2018 2nd~\cite{gao2018solution} & \cmark & 62.16 \\
2018 3rd    & \cmark & 61.70 \\       
Ours        & \cmark &\textbf{67.17} \\
\hline
\hline
Baseline (ResNeXt-152)       &     & 53.88 \\
+Class-aware Sampling                &   & 57.56 \\
+\textbf{Concurrent Softmax Loss}   &     & 58.60 \\
+\textbf{Soft-balance}              &     & 59.86 \\
+\textbf{Hybrid Training Scheduler} &     & 60.90 \\
+Other Tricks               &    & 62.34 \\
+Ensemble                  & \cmark      & \textbf{67.17} \\
\hline
\end{tabular}
\end{center}
\end{table}

With the proposed concurrent softmax, soft-balance and hybrid training scheduler,
we achieve 67.17 mAP and 4.29 points absolute improvement compared to the first place entry on the public test-challenge set last year, as detailed in Table~\ref{tab:challenge}.
We train a ResNeXt-152 FPN with multi-scale training and testing as our baseline which achieves 53.88 mAP.
After using class-aware balance, the performance is boosted to 57.56.
With the help of proposed concurrent softmax, the model achieves 58.60 mAP.
The soft-balance and the hybrid training scheduler lead to mAP gains of 1.26 and 1.04 points, respectively.
By further using other tricks including data augmentation, loss function search, and heavier head, we achieve a best single model with a mAP of 62.34.
We use ResNeXt-101, ResNeXt-152, and EfficientNet-B7 with various tricks for model ensembling.
The final mAP on Open Images public test-challenge set is 67.17.

\section{Conclusion}
In this paper, we investigate the multi-label problem and the imbalanced label distribution problem in large-scale object detection dataset 
, and introduce a simple but powerful solution.
We propose the concurrent softmax function to deal with explicit and implicit multi-label problem in both training and testing stage. Our soft-balance method together with hybrid training scheduler could effectively deal with the extremely imbalanced label distribution. 


\section{Acknowledgements}
This work was supported in part by the Major Project for New Generation of AI (No.2018AAA0100400), the National Natural Science Foundation of China (No.61836014, No.61761146004, No.61773375, No.61602481), the Key R\&D Program of Shandong Province (Major Scientific and Technological Innovation Project) (NO.2019JZZY010119), and CAS-AIR. We also thank Changbao Wang, Cunjun Yu, Guoliang Cao and Buyu Li for their precious discussion and help.


{\small
\bibliographystyle{ieee_fullname}
\bibliography{egbib}

\begin{thebibliography}{10}\itemsep=-1pt

\bibitem{akiba2018pfdet}
Takuya Akiba, Tommi Kerola, Yusuke Niitani, Toru Ogawa, Shotaro Sano, and Shuji
  Suzuki.
\newblock Pfdet: 2nd place solution to open images challenge 2018 object
  detection track.
\newblock {\em arXiv preprint arXiv:1809.00778}, 2018.

\bibitem{boutell2004learning}
Matthew~R Boutell, Jiebo Luo, Xipeng Shen, and Christopher~M Brown.
\newblock Learning multi-label scene classification.
\newblock {\em Pattern recognition}, 37(9):1757--1771, 2004.

\bibitem{cai2018cascade}
Zhaowei Cai and Nuno Vasconcelos.
\newblock Cascade r-cnn: Delving into high quality object detection.
\newblock In {\em Proceedings of the IEEE conference on computer vision and
  pattern recognition}, pages 6154--6162, 2018.

\bibitem{chawla2002smote}
Nitesh~V Chawla, Kevin~W Bowyer, Lawrence~O Hall, and W~Philip Kegelmeyer.
\newblock Smote: synthetic minority over-sampling technique.
\newblock {\em Journal of artificial intelligence research}, 16:321--357, 2002.

\bibitem{chen2019multi}
Zhao-Min Chen, Xiu-Shen Wei, Peng Wang, and Yanwen Guo.
\newblock Multi-label image recognition with graph convolutional networks.
\newblock In {\em Proceedings of the IEEE Conference on Computer Vision and
  Pattern Recognition}, pages 5177--5186, 2019.

\bibitem{cui2019class}
Yin Cui, Menglin Jia, Tsung-Yi Lin, Yang Song, and Serge Belongie.
\newblock Class-balanced loss based on effective number of samples.
\newblock In {\em Proceedings of the IEEE Conference on Computer Vision and
  Pattern Recognition}, pages 9268--9277, 2019.

\bibitem{dai2017deformable}
Jifeng Dai, Haozhi Qi, Yuwen Xiong, Yi Li, Guodong Zhang, Han Hu, and Yichen
  Wei.
\newblock Deformable convolutional networks.
\newblock In {\em Proceedings of the IEEE international conference on computer
  vision(ICCV)}, 2017.

\bibitem{deng2009imagenet}
Jia Deng, Wei Dong, Richard Socher, Li-Jia Li, Kai Li, and Li Fei-Fei.
\newblock Imagenet: A large-scale hierarchical image database.
\newblock In {\em 2009 IEEE conference on computer vision and pattern
  recognition}, pages 248--255. Ieee, 2009.

\bibitem{everingham2010pascal}
Mark Everingham, Luc Van~Gool, Christopher~KI Williams, John Winn, and Andrew
  Zisserman.
\newblock The pascal visual object classes (voc) challenge.
\newblock {\em International journal of computer vision}, 88(2):303--338, 2010.

\bibitem{gao2018solution}
Yuan Gao, Xingyuan Bu, Yang Hu, Hui Shen, Ti Bai, Xubin Li, and Shilei Wen.
\newblock Solution for large-scale hierarchical object detection datasets with
  incomplete annotation and data imbalance.
\newblock {\em arXiv preprint arXiv:1810.06208}, 2018.

\bibitem{gong2013deep}
Yunchao Gong, Yangqing Jia, Thomas Leung, Alexander Toshev, and Sergey Ioffe.
\newblock Deep convolutional ranking for multilabel image annotation.
\newblock {\em arXiv preprint arXiv:1312.4894}, 2013.

\bibitem{he2018rethinking}
Kaiming He, Ross Girshick, and Piotr Doll{\'a}r.
\newblock Rethinking imagenet pre-training.
\newblock {\em arXiv preprint arXiv:1811.08883}, 2018.

\bibitem{hu2016learning}
Hexiang Hu, Guang-Tong Zhou, Zhiwei Deng, Zicheng Liao, and Greg Mori.
\newblock Learning structured inference neural networks with label relations.
\newblock In {\em Proceedings of the IEEE Conference on Computer Vision and
  Pattern Recognition}, pages 2960--2968, 2016.

\bibitem{huang2016learning}
Chen Huang, Yining Li, Chen Change~Loy, and Xiaoou Tang.
\newblock Learning deep representation for imbalanced classification.
\newblock In {\em Proceedings of the IEEE conference on computer vision and
  pattern recognition}, pages 5375--5384, 2016.

\bibitem{krizhevsky2012imagenet}
Alex Krizhevsky, Ilya Sutskever, and Geoffrey~E Hinton.
\newblock Imagenet classification with deep convolutional neural networks.
\newblock In {\em Advances in neural information processing systems}, pages
  1097--1105, 2012.

\bibitem{kuznetsova2018open}
Alina Kuznetsova, Hassan Rom, Neil Alldrin, Jasper Uijlings, Ivan Krasin, Jordi
  Pont-Tuset, Shahab Kamali, Stefan Popov, Matteo Malloci, Tom Duerig, et~al.
\newblock The open images dataset v4: Unified image classification, object
  detection, and visual relationship detection at scale.
\newblock {\em arXiv preprint arXiv:1811.00982}, 2018.

\bibitem{li2016conditional}
Qiang Li, Maoying Qiao, Wei Bian, and Dacheng Tao.
\newblock Conditional graphical lasso for multi-label image classification.
\newblock In {\em Proceedings of the IEEE Conference on Computer Vision and
  Pattern Recognition}, pages 2977--2986, 2016.

\bibitem{li2014multi}
Xin Li, Feipeng Zhao, and Yuhong Guo.
\newblock Multi-label image classification with a probabilistic label
  enhancement model.
\newblock In {\em UAI}, volume~1, page~3, 2014.

\bibitem{li2019scale}
Yanghao Li, Yuntao Chen, Naiyan Wang, and Zhaoxiang Zhang.
\newblock Scale-aware trident networks for object detection.
\newblock {\em arXiv preprint arXiv:1901.01892}, 2019.

\bibitem{lin2017focal}
Tsung-Yi Lin, Priya Goyal, Ross Girshick, Kaiming He, and Piotr Doll{\'a}r.
\newblock Focal loss for dense object detection.
\newblock In {\em Proceedings of the IEEE international conference on computer
  vision(ICCV)}, 2017.

\bibitem{lin2014microsoft}
Tsung-Yi Lin, Michael Maire, Serge Belongie, James Hays, Pietro Perona, Deva
  Ramanan, Piotr Doll{\'a}r, and C~Lawrence Zitnick.
\newblock Microsoft coco: Common objects in context.
\newblock In {\em European conference on computer vision}, pages 740--755.
  Springer, 2014.

\bibitem{liu2016ssd}
Wei Liu, Dragomir Anguelov, Dumitru Erhan, Christian Szegedy, Scott Reed,
  Cheng-Yang Fu, and Alexander~C Berg.
\newblock Ssd: Single shot multibox detector.
\newblock In {\em Proceedings of the European conference on computer
  vision(ECCV)}, 2016.

\bibitem{lu2019grid}
Xin Lu, Buyu Li, Yuxin Yue, Quanquan Li, and Junjie Yan.
\newblock Grid r-cnn.
\newblock In {\em Proceedings of the IEEE Conference on Computer Vision and
  Pattern Recognition}, pages 7363--7372, 2019.

\bibitem{mahajan2018exploring}
Dhruv Mahajan, Ross Girshick, Vignesh Ramanathan, Kaiming He, Manohar Paluri,
  Yixuan Li, Ashwin Bharambe, and Laurens van~der Maaten.
\newblock Exploring the limits of weakly supervised pretraining.
\newblock In {\em Proceedings of the European Conference on Computer Vision
  (ECCV)}, pages 181--196, 2018.

\bibitem{niitani2019sampling}
Yusuke Niitani, Takuya Akiba, Tommi Kerola, Toru Ogawa, Shotaro Sano, and Shuji
  Suzuki.
\newblock Sampling techniques for large-scale object detection from sparsely
  annotated objects.
\newblock In {\em Proceedings of the IEEE Conference on Computer Vision and
  Pattern Recognition}, pages 6510--6518, 2019.

\bibitem{ouyang2016factors}
Wanli Ouyang, Xiaogang Wang, Cong Zhang, and Xiaokang Yang.
\newblock Factors in finetuning deep model for object detection with long-tail
  distribution.
\newblock In {\em Proceedings of the IEEE conference on computer vision and
  pattern recognition}, pages 864--873, 2016.

\bibitem{peng2019pod}
Junran Peng, Ming Sun, Zhaoxiang Zhang, Tieniu Tan, and Junjie Yan.
\newblock Pod: Practical object detection with scale-sensitive network.
\newblock In {\em Proceedings of the IEEE International Conference on Computer
  Vision}, pages 9607--9616, 2019.

\bibitem{redmon2016you}
Joseph Redmon, Santosh Divvala, Ross Girshick, and Ali Farhadi.
\newblock You only look once: Unified, real-time object detection.
\newblock In {\em Proceedings of the IEEE conference on computer vision and
  pattern recognition}, pages 779--788, 2016.

\bibitem{ren2015faster}
Shaoqing Ren, Kaiming He, Ross Girshick, and Jian Sun.
\newblock Faster r-cnn: Towards real-time object detection with region proposal
  networks.
\newblock In {\em Advances in neural information processing systems}, pages
  91--99, 2015.

\bibitem{shao2019objects365}
Shuai Shao, Zeming Li, Tianyuan Zhang, Chao Peng, Gang Yu, Xiangyu Zhang, Jing
  Li, and Jian Sun.
\newblock Objects365: A large-scale, high-quality dataset for object detection.
\newblock In {\em Proceedings of the IEEE International Conference on Computer
  Vision}, pages 8430--8439, 2019.

\bibitem{shen2016relay}
Li Shen, Zhouchen Lin, and Qingming Huang.
\newblock Relay backpropagation for effective learning of deep convolutional
  neural networks.
\newblock In {\em European conference on computer vision}, pages 467--482.
  Springer, 2016.

\bibitem{shrivastava2016training}
Abhinav Shrivastava, Abhinav Gupta, and Ross Girshick.
\newblock Training region-based object detectors with online hard example
  mining.
\newblock In {\em Proceedings of the IEEE conference on computer vision and
  pattern recognition}, pages 761--769, 2016.

\bibitem{tan2015learning}
Mingkui Tan, Qinfeng Shi, Anton van~den Hengel, Chunhua Shen, Junbin Gao,
  Fuyuan Hu, and Zhen Zhang.
\newblock Learning graph structure for multi-label image classification via
  clique generation.
\newblock In {\em Proceedings of the IEEE Conference on Computer Vision and
  Pattern Recognition}, pages 4100--4109, 2015.

\bibitem{wang2016cnn}
Jiang Wang, Yi Yang, Junhua Mao, Zhiheng Huang, Chang Huang, and Wei Xu.
\newblock Cnn-rnn: A unified framework for multi-label image classification.
\newblock In {\em Proceedings of the IEEE conference on computer vision and
  pattern recognition}, pages 2285--2294, 2016.

\bibitem{wang2017learning}
Yu-Xiong Wang, Deva Ramanan, and Martial Hebert.
\newblock Learning to model the tail.
\newblock In {\em Advances in Neural Information Processing Systems}, pages
  7029--7039, 2017.

\bibitem{wang2017multi}
Zhouxia Wang, Tianshui Chen, Guanbin Li, Ruijia Xu, and Liang Lin.
\newblock Multi-label image recognition by recurrently discovering attentional
  regions.
\newblock In {\em Proceedings of the IEEE international conference on computer
  vision}, pages 464--472, 2017.

\bibitem{wu2017adaptive}
Jian Wu, Anqian Guo, Victor~S Sheng, Pengpeng Zhao, Zhiming Cui, and Hua Li.
\newblock Adaptive low-rank multi-label active learning for image
  classification.
\newblock In {\em Proceedings of the 25th ACM international conference on
  Multimedia}, pages 1336--1344. ACM, 2017.

\bibitem{wu2018soft}
Zhe Wu, Navaneeth Bodla, Bharat Singh, Mahyar Najibi, Rama Chellappa, and
  Larry~S Davis.
\newblock Soft sampling for robust object detection.
\newblock {\em arXiv preprint arXiv:1806.06986}, 2018.

\bibitem{xie2019selftraining}
Qizhe Xie, Eduard Hovy, Minh-Thang Luong, and Quoc~V. Le.
\newblock Self-training with noisy student improves imagenet classification,
  2019.

\bibitem{zhang2018multilabel}
Junjie Zhang, Qi Wu, Chunhua Shen, Jian Zhang, and Jianfeng Lu.
\newblock Multilabel image classification with regional latent semantic
  dependencies.
\newblock {\em IEEE Transactions on Multimedia}, 20(10):2801--2813, 2018.

\bibitem{zou2018unsupervised}
Yang Zou, Zhiding Yu, BVK Vijaya~Kumar, and Jinsong Wang.
\newblock Unsupervised domain adaptation for semantic segmentation via
  class-balanced self-training.
\newblock In {\em Proceedings of the European Conference on Computer Vision
  (ECCV)}, pages 289--305, 2018.

\end{thebibliography}
}

\end{document}